\documentclass[a4paper, 10pt, conference]{mhs}      

\IEEEoverridecommandlockouts                              

\usepackage{graphics} 
\usepackage{epsfig} 
\usepackage{mathptmx} 
\usepackage{times} 
\usepackage{amsmath} 
\usepackage{amssymb}  
\usepackage{here}
\usepackage{threeparttable}
\usepackage{caption}

\title{\Large \bf
Human Preferences and Robot Constraints Aware Shared Control \\ for Smooth Follower Motion Execution
}

\author{Qibin Chen$^1$, Yaonan Zhu$^1$$^{\ast}$, Kay Hansel$^2$, Tadayoshi Aoyama$^1$, and Yasuhisa Hasegawa$^1$\\
1. Department of Micro-Nano Mechanical Science and Engineering, Nagoya University,\\
Nagoya, Aichi, 464-8603, Japan\\
2. Department of Computer Science, Technical University of Darmstadt,\\
Hochschulstr. 10, 64289 Darmstadt, Germany\\\\
\thanks{This work has been submitted to the IEEE for possible publication. Copyright may be transferred without notice, after which this version may no longer be accessible.}
\thanks{$^{\ast}$Corresponding author email: zhu@robo.mein.nagoya-u.ac.jp}
\thanks{This work was supported in part by JST Trilateral AI Research, Japan, under Grant JPMJCR20G8; and in part by JSPS KAKENHI under
Grant JP22K14222.}
\vspace{-14mm}
}

\begin{document}
	
	\maketitle
	\thispagestyle{empty}
	\pagestyle{empty}

	\begin{abstract}
    With the continuous advancement of robot teleoperation technology, shared control is used to reduce the physical and mental load of the operator in teleoperation system. 
    This paper proposes an alternating shared control framework for object grasping that considers both operator's preferences through their manual manipulation and the constraints of the follower robot. 
    The switching between manual mode and automatic mode enables the operator to intervene the task according to their wishes. 
    The generation of the grasping pose takes into account the current state of the operator's hand pose, as well as the manipulability of the robot. 
    The object grasping experiment indicates that the use of the proposed grasping pose selection strategy leads to smoother follower movements when switching from manual mode to automatic mode.

	\end{abstract}

	
	\section{Introduction}
Shared control, which employs automation to support human operation, has been widely used in teleoperation systems, since it effectively improves performance while reducing the physical and mental strain on users.
Object grasping is one of the most important teleoperation tasks and is the subject of this paper.
To address this task, previous work \cite{zhu2023shared} has developed a shared control system and demonstrated that the system improved grasping performance and reduced operator fatigue.
This shared control framework can guide the operator to an appropriate grasping pose with the best manipulability of the robot, by dynamically and continuously blending user intention and automation assistance.
This paper aims to address the problem of shared control system which does not consider human choices.
An alternating shared control system is proposed, which segregates the automation and manual aspects, empowering users to intervene at any time.
On the other hand, conventional continuous shared control maintains the combination of manual input and automation throughout the entire process.
In Maeda's work \cite{maeda2022blending}, the advantages of alternating shared control were demonstrated that the alternating way won a higher score for ease of use in subjective ratings while keeping the performance as the continuous method.
In addition, to achieve a smooth follower motion execution, the target grasping pose is selected from all candidates by taking into account not only the manipulability of the robot, but also the preferences of the operator.  

	
	\section{ Shared Control System with Human Preferences }

\subsection{Pose Remapping in Shared Control System}
The alternating shared control is realized by setting the status switching trigger on the VR controller, and processing the positional and directional gaps between the two states.
In our system, the pure teleoperation part establishes a transformation from the VR coordinate system ($\boldsymbol{p}_{hand,0}^{htc}$, $\boldsymbol{p}_{hand,k}^{htc}$, $0$ means the initial step, $k$ is for the k'th sampling step) to the robot base coordinate system ($\boldsymbol{p}_{e,k}^b$, $\boldsymbol{p}_{e,0}^b$), by calculating the relative position: $\boldsymbol{p}_{e,k}^b = \boldsymbol{p}_{e,0}^b + (\boldsymbol{p}_{hand,k}^{htc}-\boldsymbol{p}_{hand,0}^{htc})$.
For the rotational motion mapping, the absolute orientation related to the robot base frame is used as $\boldsymbol{R}_{e,k}^b = \boldsymbol{R}_{htc}^b \cdot \boldsymbol{R}_{hand,k}^{htc}$.
When the status change is triggered, the position remapping is as follows.
\begin{equation}
    \boldsymbol{p}_{e,0}^b=\boldsymbol{p}_{e,k}^b
\end{equation}
\begin{equation}
    \boldsymbol{p}_{hand,0}^{htc}=\boldsymbol{p}_{hand,k}^{htc}
\end{equation}
Meanwhile, the new absolute orientation is filtered by the spheical linear interpolation (SLERP). 
\begin{equation}
    \boldsymbol{R}_{e,k}^b=SLERP(\boldsymbol{R}_{e,k-1}^b,\boldsymbol{R}_{htc}^b \cdot \boldsymbol{R}_{hand,k}^{htc},\alpha)
    \label{eq:orientation_blending}
\end{equation}
where, $\boldsymbol{\alpha}$ is the interpolation parameter between $0$ and $1$, representing linear placement position from $\boldsymbol{R_{e,k}^b}$ to $\boldsymbol{R}_{htc}^b \cdot \boldsymbol{R}_{hand,k}^{htc}$. 

\subsection{Grasping Pose Selection Based on Human Preferences and Robot Constraints}
\begin{figure*}[t]
	\centering
	\includegraphics[width=0.85\linewidth]{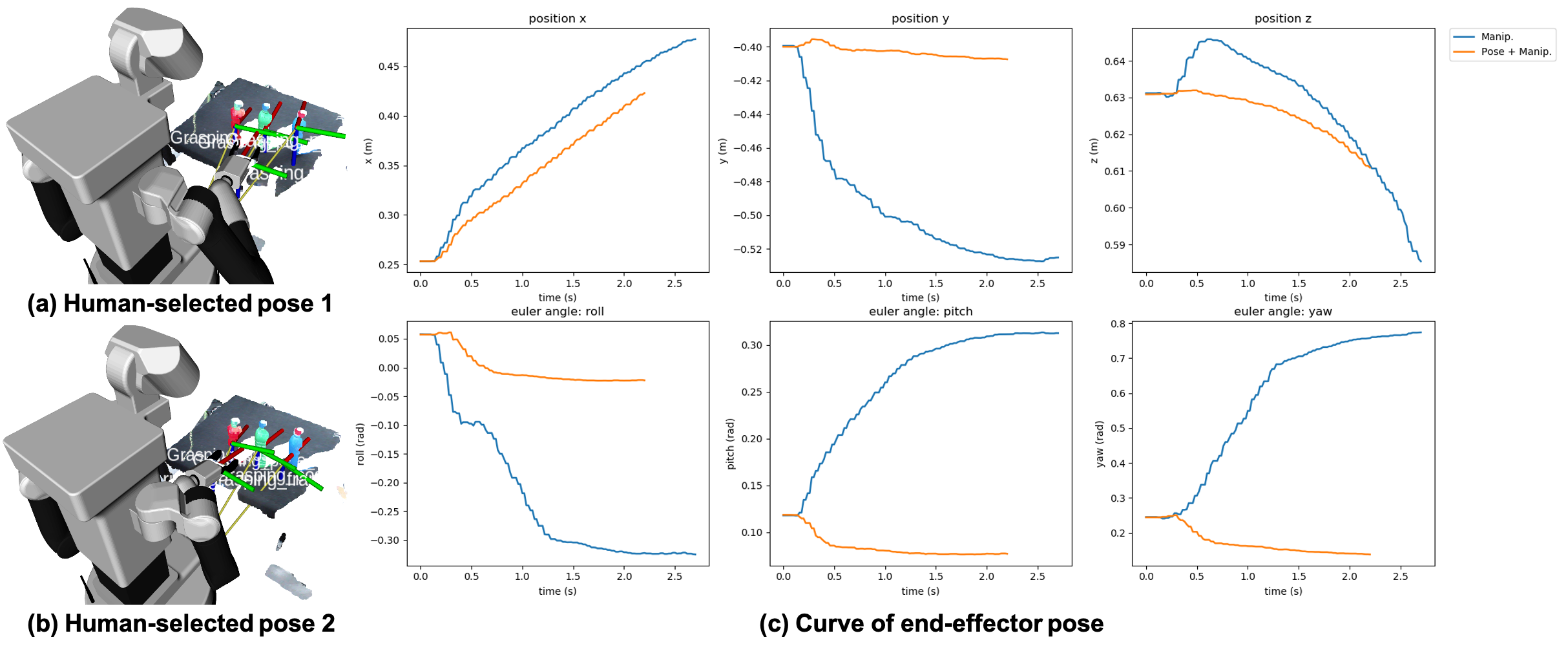}
	\caption{Target grasping poses with respect to human preferences, and end-effector motion curves with two selection strategies. 
 In (a)(b), the three-dimentional markers demonstrate the current pose of gripper and the selected grasping poses towards different objects. 
 As the user adjusts the end-effector pose from (a) pose 1 to (b) pose 2, the selected grasping poses of three objects chang accordingly.
 In (c), the method that considers both human-selected end-effector pose and robot manipulability (orange curve) makes the motion trajectories shorter and smoother than the method which considers only robot manipulability (blue curve).}
	\label{fig.1}
 \vspace{-5mm}
\end{figure*}


In \cite{zhu2023shared}, the target grasping poses were detected from multiple directions using template matching based object point cloud compensation. 
However, the final selection of the grasping pose was solely based on the best robot manipulability from the library of generated poses. 
As a result, when utilized in the alternating shared control, the user's preferences were not taken into account. 
It led to a noticeable jump in robot motion when transitioning from manual mode to automatic mode.


To enable the system to provide automatic assistance based on human manipulation outcomes, this paper proposes a feasible solution. The solution follows these steps: 
Firstly, for each object, 150 reliable generated grasping poses are stored when the object is not visually occluded by the manipulator. 
Secondly, it filters out candidates with the closest positional and directional distances in sequence. 
Thirdly, it selects the most robot-operable target grasping pose among them.
The reason is to avoid wrong grasping poses being included in the grasping pose library. 
Given the current quaternions of the end-effector $\boldsymbol{q}_{ee}$ and candidates in the grasping pose list $\boldsymbol{q}_l=[\boldsymbol{q}_1,\boldsymbol{q}_2,…,\boldsymbol{q}_i]$, the system calculates the angular absolute distance for each candidate.
\begin{equation}
    d_{a,i}=min(\lVert \boldsymbol{q}_{ee} + \boldsymbol{q}_{i} \rVert , 
    \lVert \boldsymbol{q}_{ee} - \boldsymbol{q}_{i} \rVert)
\end{equation}
The distance is the chord length of the shortest path that connects the two quaternions. 
The top 30 grasping poses closest in the orientation are updated in the candidate list.
Sequentially, the linear distance is calculated between the gripper position $\boldsymbol{p}_{ee}$ and the candidate $\boldsymbol{p}_i$ from the updated list.
\begin{equation}
    d_{l,i}=\lVert \boldsymbol{p}_{ee}-\boldsymbol{p}_i \rVert
\end{equation}
The top 6 grasping poses with the shortest distances are collected into a new list.
Then, the robot manipulabilities are calculated and the highest result is choosed as the target grasping pose. The penalized manipulability, which considers singular configurations $ S(\boldsymbol{\theta})$ and joint limits $L(\boldsymbol\theta)$ by $ M(\boldsymbol\theta) = S(\boldsymbol\theta)L(\boldsymbol\theta)$, is presented in \cite{zhu2023shared}.
Automatic approach towards the target is achieved through interpolation from the current pose to the target pose.
Hence, the grasping pose changes according to the user manipulation and the robot constraints (Fig. 1(a)(b)).
	\section {Experiment and Result}
As the improvement of manipulability on the follower side has been evaluated in our previous work \cite{zhu2023shared}, here we check the level of motion smoothness when the user preferences are carried out in the automatic mode.
The object grasping experiment is conducted in two modes: one considering only the robot constraints, and the proposed method which considers both the robot manipulability and the robot current pose. 
In each mode, the system first automatically moves to a fixed preparation pose to simulate user operation, and then completes the grasping task in automatic mode.
The Fig. 1(c) demonstrates the transformation of the robot's end-effector pose from the ready pose to the selected grasping pose.
It shows that the proposed grasping pose selection strategy reduces gripper movements, making them smoother, and achieves a shorter completion time as the execution speed is the same ($0.1m/s$).



\section{Conclusions}
In this paper, we have applied an alternating shared control that enables human intervention, enhances the comprehensibility of the robot's motion, and smoothes the gap between control mode switching. 
Positional remapping and directional spherical linear interpolation are employed to realize the intervention-possible alternating shared control.
The proposed grasping pose selection takes into account both manual operation and robot manipulability. The two factors are affected by human preferences and robot constraints, respectively.
The result of an object grasping experiment shows that the shared control system extends the operators' wishes and makes the follower's motion smooth when switching control modes.
Future work includes making different degrees of freedom have different weights in the cost function of grasping pose selection, and enabling people to fine-tune the final grasping pose.


\bibliographystyle{IEEEtran}
\bibliography{arxive}

\end{document}